\documentclass[]{bytedance_seed}

\usepackage[toc,page,header]{appendix}

\usepackage{minitoc}
\usepackage{booktabs}
\usepackage{makecell}
\usepackage{wrapfig}
\usepackage{tabularx}
\usepackage{multicol}
\usepackage{paracol}
\usepackage{adjustbox}
\usepackage[most]{tcolorbox}
\usepackage{amsmath} 
\usepackage{algorithm}
\usepackage{algpseudocode} 

\definecolor{lightgreen}{RGB}{144, 238, 144}
\definecolor{lightred}{RGB}{240, 128, 128}
\definecolor{lightyellow}{RGB}{255, 255, 50}

\newcommand{\method}{PLT }

\newcounter{globalalgorithm}

\usepackage{mdframed}
\usepackage{xcolor}
\newmdenv[
  backgroundcolor=yellow!20,
  linecolor=orange,
  linewidth=1pt,
  roundcorner=4pt,
  innertopmargin=6pt,
  innerbottommargin=6pt,
  innerleftmargin=6pt,
  innerrightmargin=6pt
]{takeawaybox}

\title{Parallel Loop Transformer for Efficient \\Test-Time Computation Scaling}

\affiliation[]{ByteDance Seed}
\contribution{Full author list in Contributions}

\abstract{

Large Language Models (LLMs) are powerful but often too slow and costly for real-world use during inference. Looped transformers save on parameters by reusing the same weights for multiple computational steps, or "loops." However, this approach has a major flaw: the loops run one after another, causing inference latency and memory requirements to increase with each added loop. This makes them impractical for fast applications.
To solve this problem, we introduce the Parallel Loop Transformer (PLT). PLT is a new architecture that delivers the performance benefits of a deep, looped model but with the low latency of a standard, non-looped model.
PLT works using two key techniques. First, Cross-Loop Parallelism (CLP) breaks the sequential dependency by computing different loops for different tokens at the same time, all within a single pass. Second, to prevent memory costs from growing, we use an Efficient Representation Enhancement strategy. This method shares the memory (KV cache) from the first loop with all other loops. It then uses a Gated Sliding-Window Attention (G-SWA) to combine this shared global information with local information, maintaining high accuracy.
Our experiments show that PLT achieves the high accuracy of a traditional looped model but with almost no extra latency or memory cost compared to a standard transformer. 

}

\date{\today}
\correspondence{Bohong Wu at \email{bohongwu@bytedance.com}, Xingyan Bin at \email{binxingyan@bytedance.com}}

\begin{document}
\maketitle



\section{Introduction}
Large Language Models (LLMs) have demonstrated remarkable capabilities across a wide range of tasks~\citep{team2023gemini,team2024gemini,liu2024deepseek,liu2024deepseekv3,li2025minimax,jiang2023mistral7b}, yet their practical deployment is often constrained by substantial computational costs during inference ~\citep{yuan2024llm,zhou2024survey,kwon2023efficient,efficientqat,qat_sl}. As models scale, the latency and memory bandwidth required for token generation become significant bottlenecks. This challenge has spurred research into architectures that can achieve high performance while maintaining inference efficiency.

\begin{figure}[tbp]
    \centering
    \begin{subfigure}[b]{0.44\textwidth}
        \centering
        \includegraphics[width=\textwidth]{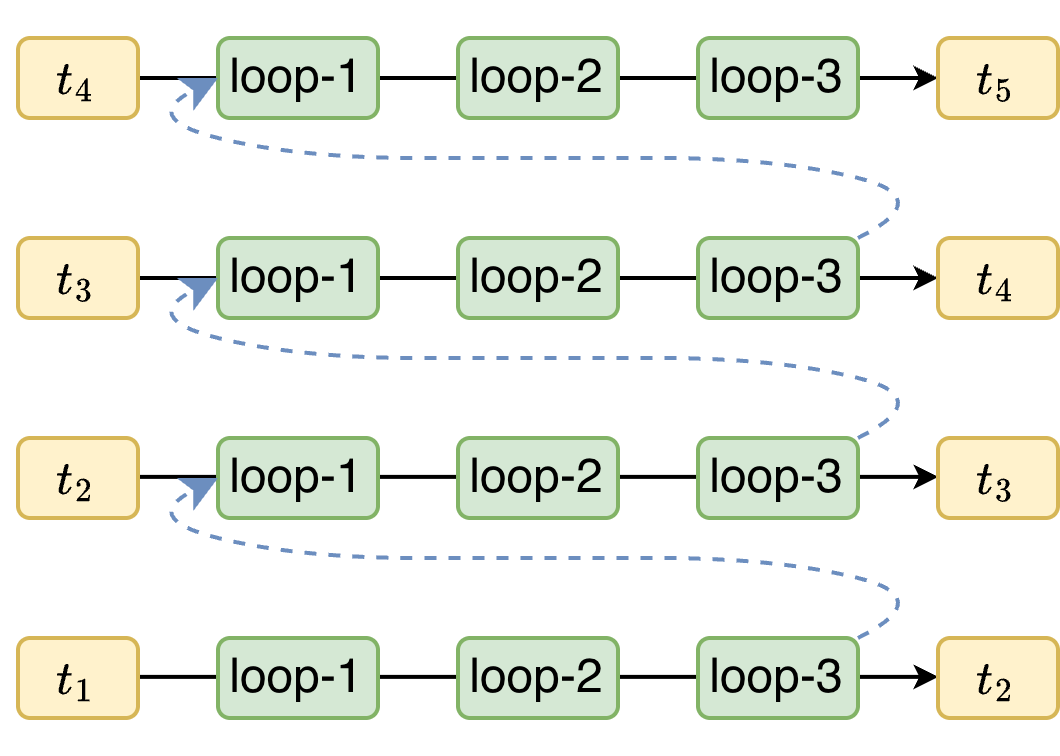}
        \caption{Inference pipeline in vanilla loop transformer}
        \label{fig:loop_trans}
    \end{subfigure}
    \hfill
    \begin{subfigure}[b]{0.44\textwidth}
        \centering
        \includegraphics[width=\textwidth]{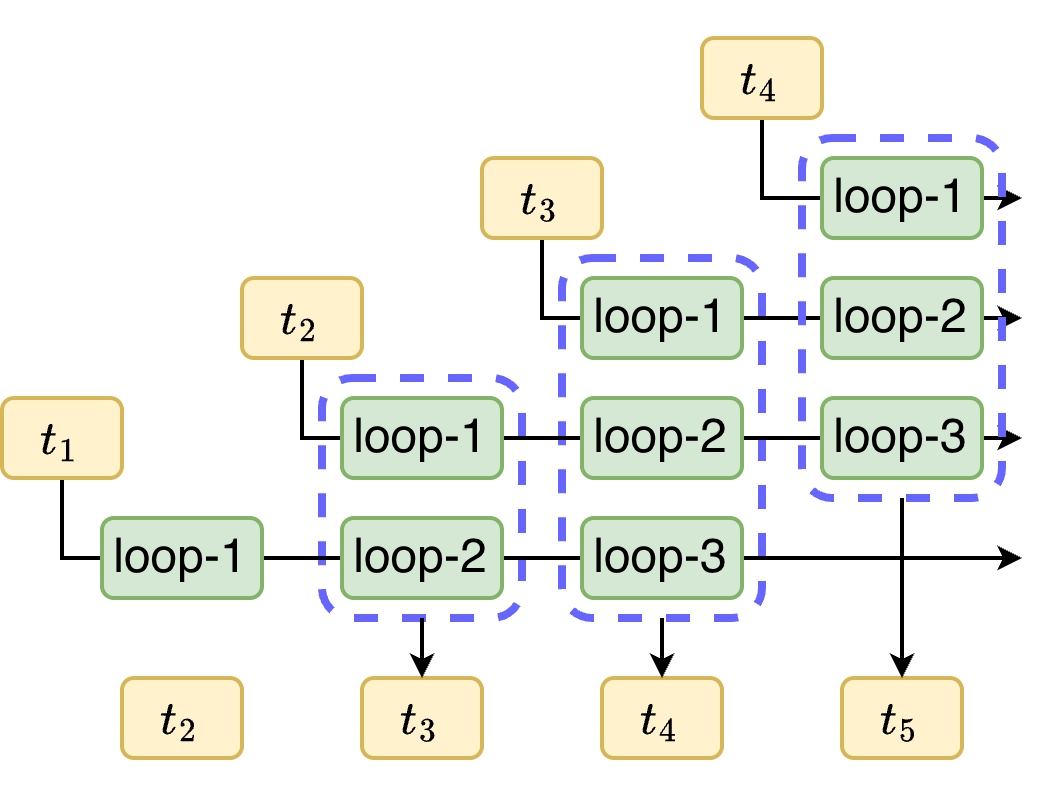}
        \caption{Inference pipeline in proposed \method.}
        \label{fig:our_infer}
    \end{subfigure}
    
    \caption{Illustration of the computation flow. (a) Vanilla loop transformer, where each loop in each token should be computed in serial manner. 
    (b) Parallel loop transformer (PLT), where transformer loops within the same {\color{blue}{blue dashed box}} can be computed in parallel.}
    \label{fig-1: PHD-v2-motivation}
\end{figure}

One promising direction is the use of looped transformers, such as the Universal Transformer~\citep{dehghani2018universal}, which reuses the same set of parameters across multiple computational steps or "loops." This weight-sharing mechanism makes them highly parameter-efficient, enabling them to achieve greater effective depth and stronger performance on complex reasoning tasks without increasing the model's storage footprint~\citep{fan2024looped,yang2023looped,saunshi2025reasoning,tang2024quest,mohtashami2023cotformer,tack2025llm,zelikman2024quiet,hao2024training}. However, this parameter efficiency comes at a severe cost: in their vanilla implementation, the loops are executed in a strictly sequential manner as shown in Figure~\ref{fig:loop_trans}. Such sequential dependency means that per-token compute, wall-clock latency, and KV-cache size all scale linearly, $\mathcal{O}(L)$, with the number of loops ($L$). This scaling bottleneck renders vanilla looped transformers impractical for latency-sensitive applications, effectively negating their primary advantages.

To resolve this tension between effective depth and inference speed, we introduce the \textbf{Parallel Loop Transformer (PLT)}. PLT is a novel architecture designed to unlock the performance benefits of deep, looped computation while maintaining the approximate inference latency of a standard, non-looped transformer. As shown in Figure~\ref{fig:our_infer}, the core principle of PLT is to break the sequential loop dependency by parallelizing the computation of different loops across different tokens.

Our approach consists of two key components. First, we introduce \textbf{Cross-Loop Parallelism (CLP)}, a technique that reformulates the training and inference pipelines. During decoding, CLP executes the $l$-th loop for the current token $t_i$ concurrently with the $(l+1)$-th loop for the previous token $t_{i-1}$ and so on, all within a single forward pass. This overlapping computation effectively collapses the $L$ sequential steps into one, decoupling the model's effective depth from its wall-clock latency. Second, to address the $\mathcal{O}(L)$ growth in KV cache, we propose an \textbf{Efficient Representation Enhancement} strategy. This strategy shares the KV cache from the first loop with all subsequent loops, reducing the KV cache memory footprint back to the level of Loop times as 1. To preserve high accuracy, we augment this global shared representation with a local context mechanism, using a gated sliding-window attention (SWA) in non-first loops.

Our contributions are as follows:
\begin{itemize}
    \item We propose the Parallel Loop Transformer (PLT), an architecture that, to our knowledge, is the first to successfully parallelize the computation of looped transformers to achieve scalable test-time computation with negligible latency overhead.
    \item We introduce Cross-Loop Parallelism (CLP) and an Efficient Representation Enhancement technique (KV-cache sharing with gated SWA) to overcome the critical latency and memory bottlenecks of traditional looped models.
    \item We demonstrate through extensive experiments on both in-house and open-source models that PLT significantly outperforms vanilla transformer baselines in accuracy while adding minimal latency.
    \item We show that PLT is far more efficient than vanilla looped transformers, and can even enable a shallower, more efficient PLT model (e.g., 1.7B activated parameters) to achieve superior performance and lower latency than a much larger vanilla model (e.g., 2.5B activated parameters).
\end{itemize}

\section{Method}

\subsection{Preliminaries: Vanilla Loop Transformer}\label{sec:preliminary}

We consider a \textbf{vanilla Loop Transformer}~\citep{dehghani2018universal} with $L$ loops.
Let $T = (t_1, t_2, \dots, t_n)$ be a token sequence, $\mathbf{E} = (\mathbf{e}_1, \mathbf{e}_2, \dots, \mathbf{e}_n)$ is the token sequence after embedding. For token index $i \in \{1,\dots,n\}$ and loop index $l \in \{1,\dots,L\}$, let $h_i^{(l)}$ denote the hidden state at position $i$ after $l$ forward loops, with $h_i^{(0)}$ the initial state.

The computation flow for $i$-th token in vanilla loop transformer is as follow:
\begin{equation}
h_i^{(0)} = t_i, 
\qquad
h_i^{(l)} = f^{(l)}\!\left(h_i^{(l-1)}\right), \quad l = 1,\dots,L,
\label{eq:loop}
\end{equation}
where $f^{(l)}$ denotes the $l$-th forward loop. Finally, $h_i^{(L)}$ feeds into the classifier head to predict the $(i+1)$-th token.

\textbf{Challenge in vanilla loop transformer.}
In a vanilla loop transformer, loops run strictly sequentially, so per-token compute, wall-clock latency, and KV-cache size scale as $\mathcal{O}(L)$ with the number of loops $L$ as shown in Table~\ref{tab:efficiency}.
While weight sharing makes it parameter-efficient (it achieves greater effective depth with fewer stored weights), it does not reduce  latency.
Therefore, it mainly helps under equal-parameter comparisons; under equal-latency budgets—typical in practical inference—the vanilla loop transformer offers no inherent advantage and can be worse due to longer decode paths, higher memory-bandwidth pressure, and larger KV caches.

\subsection{Parallel Loop Transformer}
\textbf{Motivation.}
Vanilla loop transformers are parameter-efficient but rely on strictly serial computation per token, which increases the forward times and the KV-cache footprint. We aim to keep the parameter savings while shortening the forward times by enabling parallelism across loops and tokens without breaking causality.

As shown in Figure~\ref{fig:our_infer}, we convert the vanilla sequential loop transformer into a parallel loop transformer that executes $L$ loops in parallel. The design has two key components:
(i) cross-loop parallelism (Sec.~\ref{sec:displaced_parallism}) for parallel loop inference;
(ii) efficient representation enhancement (Sec.~\ref{sec:local_info}) to improve accuracy with light time and KV cache overhead.

\subsubsection{Cross‑Loop Parallelism}~\label{sec:displaced_parallism} \label{sec:strectch_activation}

\begin{figure}[tbp]
    \centering
    \includegraphics[width=0.95\textwidth]{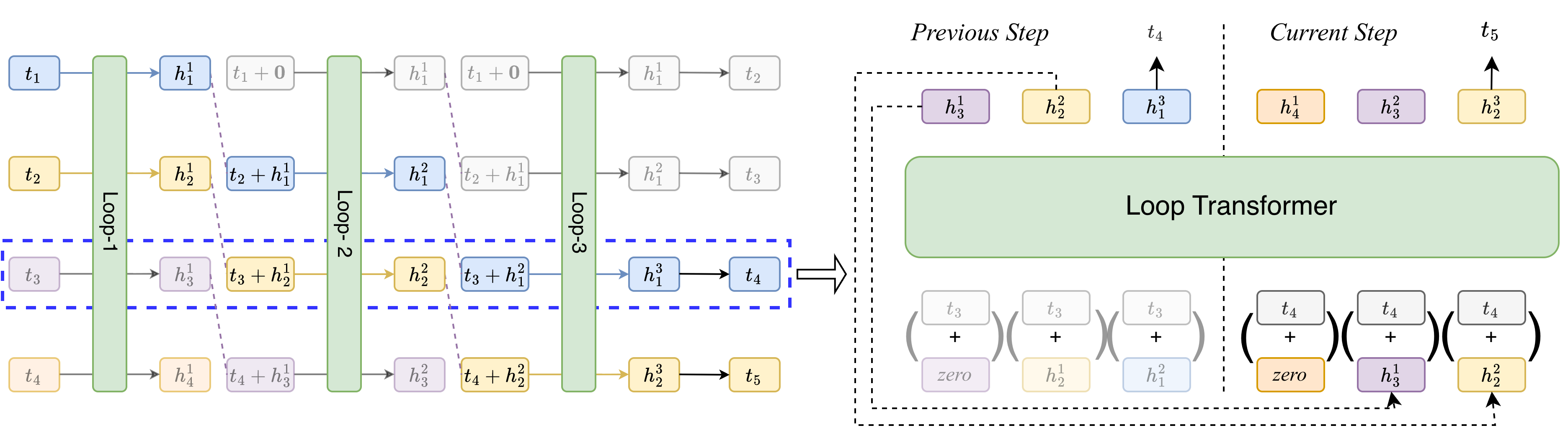}
\caption{
Training and inference pipeline of \method with loop count $L{=}3$. 
\textbf{Training (Left):} Same Colored boxes trace how input tokens traverse the loops to predict their targets (e.g., token $T_1$ passes three loops to predict $T_4$, consistent with Figure~\ref{fig:our_infer}). Training is parallel along the \emph{token} dimension and serial along the \emph{loop} dimension. 
\textbf{Inference (Right):} Parallelized forward pass of \method when decoding $T_4$ and $T_5$ in a Loop Transformer with $L{=}3$. Because there are no horizontal (same-step, cross-loop) activation dependencies during training, computations within the same step (each row; see the blue dashed box) run in parallel during decoding.
}

    \label{fig:loop_train_sharekv}
\end{figure}

\stepcounter{globalalgorithm}
\begin{algorithm}[H]
        \caption{Decoding of PLT with loop times as 3}\label{alg:decoding}
        \begin{algorithmic}[1]
        \Require
        \Statex Loop times $L=3$
        \Statex Transformer block function $f$;
        \Statex Input sequence $T = (t_1, \dots, t_n)$;
        \Statex Maximum new tokens $M$.
        
        \State $K_{share}$, $V_{share}$, $h_{n}^{1}$, $h_{n-1}^{2}$, $h_{n-2}^{3} \gets f(T)$ \Comment{Prefilling}
        
        \State $\text{logits} \gets \text{ClassifierHead}(h_{k-2}^{3})$
        \State $t_{k+1} \gets \text{argmax}(\text{logits})$ \Comment{Predict the next token}

        \For{$i = n+1$ \textbf{to} $n+M-1$}
            \State $\mathbf{e}_i = \text{Embedding}(t_i)$ 
            
        
            \State $\mathbf{B} \gets {B_0, B_1, B_2}\gets{\mathbf{e}_i, \mathbf{e}_i+h_{i-1}^{1}, \mathbf{e}_i+h_{i-2}^{2}}$
        
            \State $K_{share},V_{share},h_{i}^{1},h_{i-1}^{2},h_{i-2}^{3} \gets f(\mathbf{B}, K_{share},V_{share})$

            \State $\text{logits} \gets \text{ClassifierHead}(h_{i-2}^{3})$
            \State $t_{i+1} \gets \text{argmax}(\text{logits})$
        \EndFor
\end{algorithmic}
\end{algorithm}

\begin{paracol}{2}
    \stepcounter{globalalgorithm}
    \begin{algorithm}[H]
        \caption{Training for PLT}\label{alg:training}
         \begin{algorithmic}[1]
             \Require
                \Statex \quad Loop count $L$
                \Statex \quad Transformer block function $f$
                \Statex \quad Classification head $\text{ClassifierHead}(\cdot)$
                \Statex \quad Input token sequence $T = (t_1, \dots, t_n)$

                \Statex
                \State $\mathbf{E} \gets \text{Embedding}(T)$   
                \State $K_{\text{share}}, V_{\text{share}}, \mathbf{H}^{(1)} \gets f\!\left(\mathbf{E}\right)$  
        
                \For{$i = 2$ \textbf{to} $L$} 
                    \State $\mathbf{H}^{(i-1)} \gets \text{concat}(\mathbf{0},\mathbf{H}^{(i-1)}[:-1])$
                    \State $\mathbf{B} \gets \mathbf{E} + \mathbf{H}^{(i-1)}$ 
                    \State $\mathbf{H}^{(i)} \gets f\!\left(\mathbf{B};\, K_{\text{share}}, V_{\text{share}}\right)$  
                \EndFor
        
                \State $\text{logits} \gets \text{ClassifierHead}\!\left(\mathbf{H}^{(L)}\right)$ 
                \State $loss = \text{CrossEntropy}(\text{logits}, T)$ 
                \State \Return $loss$
         \end{algorithmic}
    \end{algorithm}
    \switchcolumn
    
    \stepcounter{globalalgorithm}
    \stepcounter{globalalgorithm}
    \begin{algorithm}[H]
        \caption{Efficient representation enhancement for non-first loop in PLT}\label{alg:gatedswa}
         \begin{algorithmic}[1]
             \Require
                \Statex \quad Input hidden state $\mathbf{H}$
                \Statex \quad QKV linear layer $f_{qkv}$, Output linear layer $f_{o}$
                \Statex \quad Gate linear layer $f_{gate}$
               
                \Statex \quad shared Key-Value cache from first Loop $(K_{\mathrm{share}}, V_{\mathrm{share}})$
                \Statex \quad window size $w$ of sliding window attention (SWA)
        
                \State $Q, K, V = f_{qkv}(\mathbf{H})$
                \State $y_{\mathrm{global}} = \mathrm{Attn}(Q,\ K_{\mathrm{share}}, V_{\mathrm{share}})$
                \State $y_{\mathrm{local}} = \mathrm{SWA}(Q,\ K, V, w)$
                \State $g = \text{Sigmoid}\!\big(f_{gate}(Q))$ 
                \State $\tilde{y} = g \odot y_{\mathrm{local}} + (1{-}g) \odot y_{\mathrm{global}}.$
                \State $o = f_{o}(\tilde{y})$
        
                \State \Return $o$
         \end{algorithmic}
    \end{algorithm}
\end{paracol}

We introduce \emph{Cross‑Loop Parallelism} (CLP): overlapping later-loop computation on earlier tokens with earlier-loop computation on later tokens. As shown in Figure~\ref{fig:our_infer}, taking decoding token $t_4$ with 3 loops as example, first loop on token $t_3$, second loop on token $t_2$ and third loop on token $t_1$ are executed simultaneously, and similar cross-loop parallelism are executed by the following tokens. The detailed inference and training pipeline of this method are proposed as follows:

\textbf{Training.} The training flow of PLT is illustrated in Figure~\ref{fig:loop_train_sharekv} and Algorithm~\ref{alg:training}. In the first loop, PLT matches a vanilla Transformer: it feeds the token embeddings $\mathbf{H}^{(0)} = \mathbf{E} = (\mathbf{e}_1, \mathbf{e}_2, \dots, \mathbf{e}_n)$ into the model and obtains the last-layer hidden states $\mathbf{H}^{(1)} = (\mathbf{h}_1^{(1)}, \mathbf{h}_2^{(1)}, \dots, \mathbf{h}_n^{(1)})$.
Before the second loop, PLT shifts $\mathbf{H}^{(1)}$ to the right by one position, from $(\mathbf{h}_1^{(1)}, \dots, \mathbf{h}_n^{(1)})$ to $(\mathbf{0}, \mathbf{h}_1^{(1)}, \dots, \mathbf{h}_{n-1}^{(1)})$, and then adds back the original embeddings: $\mathbf{B} = \mathbf{E} + \text{shift}(\mathbf{H}^{(1)})$. This shift removes direct dependence between the states with the same index across consecutive loops, which enables parallel processing during decoding; we refer to this as \textbf{cross-loop parallelism}. 
PLT repeats this process for $i = 2, \dots, L$: at each loop, it right-shifts the previous loop’s states by one position, adds the embeddings, and applies the transformer while reusing the shared $K_{\text{share}}$ and $V_{\text{share}}$ from the first loop.

\textbf{Inference.} As shown in Figure~\ref{fig:loop_train_sharekv} (right), \textbf{cross-loop parallelism} in PLT processes $L$ tokens with a single forward pass. This parallel design leverages the memory-bound nature~\cite{unveiled} of LLM decoding: adding parallel test-time computation FLOPs improves accuracy, while the extra decoding latency is negligible. Algorithm~\ref{alg:decoding} illustrates the decoding flow of PLT for $L{=}3$. At decoding step $i$, we construct a displaced micro-batch $\mathbf{B}{=}\{B_0, B_1, B_2\}$, where $B_0$ is the first loop of token $i$, $B_1$ is the second loop of token $i{-}1$, and $B_2$ is the third loop of token $i{-}2$. We then predict token $(i{+}1)$ from the final loop state $h_{i-2}^{3}$. Compared with a vanilla loop transformer (which applies $L$ sequential passes per token), PLT performs the same logical compute with one parallel pass per token. Compared with a standard non-loop decoder, it adds FLOPs that improve accuracy, yet the latency increase is negligible because decoding is memory-bound~\cite{unveiled}.

\subsubsection{Efficient Representation Enhancement}\label{sec:local_info}
Cross-loop parallelism addresses the inference-time scaling with loop count $L$ in a vanilla loop transformer, but it still incurs an $L$-fold KV-cache memory cost, which limits long-context use. We introduce \textbf{efficient representation enhancement} with two components: (i) first-loop KV-cache sharing to provide a single global representation, and (ii) gated sliding-window attention to strengthen local context. Details follow.

\textbf{KV-cache sharing from the first loop.} In a standard loop design, each loop maintains its own KV cache, so memory grows linearly with $L$. To reduce KV memory, we share the first loop's KV cache with all later loops. As shown in Algorithm~\ref{alg:gatedswa}, non-first loops keep their private queries but perform global attention on $K_{\mathrm{share}}$ and $V_{\mathrm{share}}$ from the first loop. Thus, only one global KV cache needs to be stored. This design preserves global information for non-first loops and removes the $L$-dependent KV-cache growth.

\textbf{Gated sliding-window attention (G-SWA) in non-first loops.} To further enhance local information on top of the shared global representation, non-first loops apply sliding-window attention over their private $Q$, $K$, and $V$ with window size $w$. In our experiments, we set $w{=}64$, and it does not increase with the overall sequence length. As shown in Lines~3--5 of Algorithm~\ref{alg:gatedswa}, we then fuse the outputs of SWA (local) $y_{\mathrm{local}}$ and full attention on the shared KV (global) $y_{\mathrm{global}}$ using a sigmoid gate:
\begin{equation}
g \gets \mathrm{Sigmoid}\!\big(f_{\mathrm{gate}}(Q)\big), \quad
\tilde{y} \gets g \odot y_{\mathrm{local}} + (1{-}g) \odot y_{\mathrm{global}}.
\end{equation}
The gate linear layer $f_{\mathrm{gate}}$ is head-wise with a scalar output per head, so the added parameters and computation are negligible. In addition, due to SWA, non-first loops only cache the most recent $w$ KV entries during decoding. Overall, gated sliding-window attention makes the proposed PLT more KV-cache efficient than a vanilla loop transformer, since non-first loops store only $w$ recent KV entries, while accuracy is maintained by combining both global and local information.


\begin{table}[tp]
    \centering
    \renewcommand{\arraystretch}{0.4}
\caption{
Complexity comparison of the vanilla Transformer, a looped Transformer, and the proposed \textit{PLT}. 
The vanilla Transformer serves as the baseline with parameter count $P$, per-token compute cost $C$, and single-token decoding latency $t$ under a memory-bound setting. 
The attention KV cache scales as $\mathcal{O}(nd)$, where $n$ is the sequence length and $d$ is the embedding dimension. 
Here, $L$ denotes the loop count and $w$ the sliding-window size used by \textit{SWA} (sliding-window attention).
}\label{tab:efficiency}
    \small
    \begin{tabular}{l|ccccc}
    \toprule
    & \textit{Loop Times} & \textit{Param.} & \textit{Compute.} & \textit{KV cache} & \textit{Decoding latency} \\
    \midrule
    (1)Vanilla transformer & $1$ & $P$ & $C$ & $\mathcal{O}(nd)$ & $t$ \\
    (2)Vanilla loop transformer & $L$ & $P$ & $LC$ & $\mathcal{O}(\textcolor{red}{L}nd)$ & $\textcolor{red}{L}t$ \\
    (3)\textit{Loop}+\textit{CLP} & $L$ & $P$ & $LC$ & $\mathcal{O}(\textcolor{red}{L}nd)$ & $\sim t$ \\
    (4)\textit{Loop}+\textit{CLP}+\textit{KV share} & $L$ & $P$ & $LC$ & $\mathcal{O}(nd)$ & $\sim t$ \\
    (5)\textit{Loop}+\textit{CLP}+\textit{KV share}+\textit{G-SWA} & $L$ & $P$ & $LC$ & $\mathcal{O}(nd + (\textcolor{red}{L}-1)wd)$ & $\sim t$ \\
    \bottomrule
    \end{tabular}
    \label{tab:exp_efficiency_analysis}
\end{table}

\subsection{Inference Efficiency Analysis}
\label{sec:inference-efficiency}
As shown in Table~\ref{tab:efficiency}, we analyze per-token inference in a memory-bound setting. The vanilla Transformer (baseline) has parameter count $P$, per-token compute $C$, KV cache $\mathcal{O}(nd)$, and decoding latency $t$. Introducing $L$ loops increases compute to $LC$ and, if each loop keeps its own cache, increases memory to $\mathcal{O}(Lnd)$; because the loops run sequentially, latency grows to $Lt$. With the proposed cross-loop parallelism, the loop iterations run concurrently, so latency is $\approx t$, compute remains $LC$, and the KV cache remains $\mathcal{O}(Lnd)$. A KV-cache sharing variant reuses the cache of the first loop and does not store separate caches for the non-first loops, which reduces the KV cache to $\mathcal{O}(nd)$ at the cost of a small accuracy drop. Finally, rather than using full-context attention or disabling attention in the non-first loops, we apply sliding-window attention of size $w$ in those loops, which yields a total KV cache of $\mathcal{O}\!\big(nd + (L-1)wd\big)$. Since typically $w \ll L$, the extra KV cache and compute overhead is small. Overall, {PLT} maintains near-baseline decoding latency and avoids the $L$-fold growth of the KV cache.


\section{Experiments}
We organize our experiments in two parts. Section~\ref{exp:same_act_inhouse} provides a comprehensive comparison of \method against a vanilla Transformer baseline under identical parameter settings, demonstrating the contribution of each component introduced in \method and examining scalability with respect to the number of parallelized loops. Section~\ref{exp:inhouse_2b5} evaluates inference efficiency at matched accuracy, showing that \method achieves performance comparable to the vanilla Transformer with fewer parameters while delivering significantly improved inference efficiency.


\subsection{Accuracy Comparisons under the Same Parameters} \label{exp:same_act_inhouse}

\subsubsection{Settings} \label{sec:training setting}

\textbf{Training recipe.} In this section, we compare Seed-MoE models with a vanilla looped Transformer and variants of \method. We add each component—cross-loop parallelism, KV-cache sharing, and gated sliding-window attention (G-SWA)—to the vanilla looped Transformer one by one to assess the effect of each component. For PLT-related hyperparameters, we set the G-SWA window size $w$ to 64 and vary the PLT loop count $L \in \{2,3\}$ to study scalability. We train 680M/13B MoE (680M activated parameters with 13B total parameters) models on 150B high-quality tokens. Additional details are withheld due to confidentiality; a more detailed open-source training configuration appears in Section~\ref{exp:part1}.

\textbf{Accuracy evaluation recipe.} We evaluate accuracy on the following open-source benchmarks: MMLU~\citep{hendrycks2020measuring}, CEval~\citep{huang2023c}, AGIEval (AGI.)~\citep{zhong2024agieval}, MMLU-Pro (M. Pro)~\citep{wang2024mmlu}, BBH~\citep{suzgun2023challenging}, DROP~\citep{dua2019drop}, MBPP~\citep{austin2021program}, HumanEval (H.Eval)~\citep{chen2021evaluating}, MATH~\citep{hendrycks2021measuring}, and GSM8k~\citep{cobbe2021training}.

\textbf{Inference efficiency evaluation recipe.} We use FP8 self-attention ~\citep{shah2024flashattention} and W4A8 quantization of linear layers~\citep{hu2025liquidgemm} to simulate real serving. To evaluate efficiency under both low- and high-throughput scenarios, we set the prefill context length to 5000 and vary the inference batch size in $[4,8,16,32,64]$, which shifts decoding from low-throughput to high-throughput scenarios. We report per-token decoding latency, averaged over five independent runs that each decode 256 tokens on one single GPU.

\begin{table}[tp]
    \centering
    \caption{
    Performance evaluation of in-house Seed-MoE and \textit{PLT} with the same number of activated parameters. Lat. and KV cache represents decoding latency (ms) and KV cache memory overhead under batch size as 4.
    }
    \scriptsize
    \begin{adjustbox}{max width=\textwidth}
    \begin{tabularx}{\textwidth}{l|XXXXXXXXXX|X|XX}
    \toprule
     & \tiny{MMLU} & \tiny{CEval} & \tiny{M.PRO} & \tiny{AGI.} & \tiny{BBH} & \tiny{DROP} & \tiny{GSM.} & \tiny{H.Eval} & \tiny{MBPP} & \tiny{TQA} & \tiny{Avg.} & \tiny{Lat.$_1$} & \tiny{kv cache} \\
    \midrule
    (1)Seed-MoE 680M/13B & 54.0 & 53.0 & 22.5 & 31.5 & 36.8 & 30.1 & 22.6 & 25.0 & 34.1 & 37.4 & 34.7 & \colorbox{lightgreen}{4.8} & 280M \\
    (2)+\textit{loop-2}  &  59.1 & 61.6 & 26.0 & 37.7 & 44.2 & 36.8 & 29.0 & 25.0 & 35.7 & 41.6 & 39.7 & \colorbox{lightred}{9.4} & 560M \\
    (3)+\textit{loop-2}+\textit{CLP}  & 59.7 & 60.6 & 28.6 & 36.6 & 38.9 & 36.8 & 34.6 & 25.6 & 38.1 & 36.3 & 39.6 & \colorbox{lightgreen}{5.9} & 560M \\
    (4)+\textit{loop-2}+\textit{CLP}+\textit{KV share}&  55.7 & 57.2 & 22.1 & 33.4 & 35.3 & 33.5 & 27.3 & 23.8 & 34.4 & 39.1 & 36.2 & \colorbox{lightgreen}{4.8} & 280M  \\
    (5)+\textit{loop-2}+\textit{CLP}+\textit{KV share}+\textit{G-SWA}(\textit{PLT}-2) &  59.6 & 58.9 & 27.0 & 34.9 & 41.6 & 36.4 & 33.6 & 26.8 & 37.8 & 40.0 & 39.7 & \colorbox{lightgreen}{4.9} & 284M  \\
    (6)+\textit{loop-3}+\textit{CLP}+\textit{KV share}+\textit{G-SWA}(\textit{PLT}-3) &  62.5 & 61.4 & 27.1 & 35.7 & 40.3 & 41.7 & 39.3 & 23.8 & 34.4 & 41.3 & \emph{40.8} & \colorbox{lightgreen}{5.0} & 287M\\
    \bottomrule
    \end{tabularx}
    \end{adjustbox}
    \label{tab:main_exp_inhouse}
\end{table}

\subsubsection{Results and Analysis}

\begin{table}[tp]
    \centering
    \caption{
    Inference efficiency evaluation of PLT variants across different batch size.
    }
    \scriptsize
    \begin{adjustbox}{max width=\textwidth}
    \begin{tabular}{l|rrrrr}
    \toprule
     & bs{=}4 & bs{=}8 & bs{=}16 & bs{=}32 & bs{=}64 \\
    \midrule
    (1)Seed-MoE 680M/13B & 4.8(1.00$\times$) & 5.6(1.00$\times$) & 6.7(1.00$\times$) & 8.1(1.00$\times$) & 10.9(1.00$\times$)  \\
    (2)+\textit{loop-2} & 9.4(1.96$\times$) & 11.1(1.98$\times$) & 13.2(1.97$\times$) & 16.1(1.99$\times$) & 21.4(1.96$\times$)  \\
    (3)+\textit{loop-2}+\textit{CLP}  & 5.9(1.23$\times$) & 6.9(1.23$\times$) & 8.5(1.27$\times$) & 10.9(1.35$\times$) & 16.4(1.50$\times$) \\
    (4)+\textit{loop-2}+\textit{CLP}+\textit{KV share} & 4.8(1.00$\times$) & 5.6(1.00$\times$) & 6.8(1.01$\times$) & 8.3(1.02$\times$) & 11.1(1.02$\times$)  \\
    (5)+\textit{loop-2}+\textit{CLP}+\textit{KV share}+\textit{G-SWA}(\textit{PLT}-2) & 4.9(1.02$\times$) & 5.7(1.02$\times$) & 6.9(1.03$\times$) & 8.6(1.06$\times$) & 11.3(1.04$\times$) \\
    \bottomrule
    \end{tabular}
    \end{adjustbox}
    \label{tab:inhouse_efficiency}
\end{table}

Table~\ref{tab:main_exp_inhouse} reports the performance of vanilla transformers, loop transformers, and variants of \method. We summarize and explain the main findings below.

\textbf{Observation 1: Cross-loop parallelism (CLP) preserves accuracy and reduces latency.}
A loop transformer with two loops (row (2)) raises average accuracy by +5.0 points over the vanilla model (34.7$\rightarrow$39.7), but also increases latency and memory (4.8$\rightarrow$9.4\,ms, +96\%; 280M$\rightarrow$560M, +100\%). Adding \textit{CLP} at the same loop count (row (3)) keeps the accuracy nearly unchanged (39.7$\rightarrow$39.6, $-0.1$), while it cuts latency from 9.4\,ms to 5.9\,ms ($-37$\%) by parallelizing loop computation. Compared to the vanilla transformer, CLP keeps most of the accuracy gain (+4.9 points) with only a modest latency factor (4.8$\rightarrow$5.9\,ms, $\approx$1.23$\times$) and no extra KV-cache versus the naive loop (still 560M). Overall, CLP removes the sequential latency cost of looping while keeping the accuracy benefit.

\textbf{Observation 2: Efficient representation enhancement reduces KV-cache with minimal accuracy loss.}
Efficient representation enhancement includes two parts: KV-cache sharing and gated sliding-window attention (G-SWA). \emph{KV-cache sharing} (row (4)) removes the extra KV-cache footprint from looping (560M$\rightarrow$280M, $-50$\%) and also reduces latency (5.9$\rightarrow$4.8\,ms, $-19$\%) because of less KV-cache loading time during decoding, but also lowers average accuracy benefit by 3.4 points (39.6$\rightarrow$36.2) because each loop no longer holds its own dedicated KV. Furthermore,  \emph{Adding G-SWA} (row (5)) restores per-loop specificity using a local window, raising accuracy from 36.2 to 39.7 (+3.5) while adding only 1.4\% KV-cache overhead (280M$\rightarrow$284M) and nearly no latency penalty (4.8$\rightarrow$4.9\,ms).  Therefore, KV-cache sharing solves the memory blow-up; G-SWA recovers accuracy at negligible cost.

\textbf{Observation 3: \method (\textit{PLT}) delivers loop-level accuracy with near-vanilla efficiency and scales well.}
With two loops, \textit{PLT}-2 (row (5)) matches the accuracy of the naive loop transformer (39.7) but keeps efficiency close to the vanilla model (latency 4.8$\rightarrow$4.9\,ms, +2\%; KV 280M$\rightarrow$284M, +1.4\%). Scaling to \textit{PLT}-3 (row (6)) further improves average accuracy to 40.8 (+1.1 over \textit{PLT}-2) with only a small latency increase (4.8$\rightarrow$4.9\,ms, +2\%) and a minor KV change (284M$\rightarrow$287M, +1.1\%). \method decouples latency and memory from the loop count, so it preserves the accuracy gains of looping while keeping inference overhead close to vanilla, and it scales smoothly with more loops.

\subsubsection{Inference Efficiency}
Table~\ref{tab:inhouse_efficiency} shows that \method (\textit{PLT}) improves latency in both low-throughput and high-throughput regimes while outperforming the vanilla loop transformer. In the low-throughput small-batch setting ($\mathrm{bs}\in\{4,8\}$),  adding KV sharing on top of CLP (row (3)$\rightarrow$(4)) reduces latency from $5.9\!\rightarrow\!4.8$\,ms ($-19\%$) at $\mathrm{bs}=4$ and $6.9\!\rightarrow\!5.6$\,ms ($-19\%$) at $\mathrm{bs}=8$, and \textit{PLT}-2 (row (5)) stays near vanilla (row (1)) at $1.02\times$ ($4.8\!\rightarrow\!4.9$\,ms; $5.6\!\rightarrow\!5.7$\,ms) while G-SWA recovers accuracy with negligible latency change. In the high-throughput large-batch setting ($\mathrm{bs}\in\{32,64\}$), CLP’s concurrency is the main driver: relative to the vanilla loop transformer (row (2)), \textit{PLT}-2 cuts per-token latency by $47\%$ at $\mathrm{bs}=32$ ($16.1\!\rightarrow\!8.6$\,ms) and $47\%$ at $\mathrm{bs}=64$ ($21.4\!\rightarrow\!11.3$\,ms), with residual overhead vs. vanilla limited to $1.06\times$ ($8.1\!\rightarrow\!8.6$\,ms) and $1.04\times$ ($10.9\!\rightarrow\!11.3$\,ms), respectively. Overall, \textit{PLT}-2 reduces latency by $47\%$ compared to naive looping across batch sizes and remains within $4$–$6\%$ of vanilla for practical serving ($\mathrm{bs}\ge 32$).

\subsection{Latency Comparisons under Same Accuracy} \label{exp:inhouse_2b5}

We scale the $PLT$ model size to match the accuracy of the vanilla Seed-MoE baseline while improving inference efficiency.

\textbf{Training recipe.} The baseline is an in-house 2.5B/60B Seed-MoE model trained on 1T tokens. For $PLT$, we use a shallower model by setting the number of layers to two-thirds of the baseline, yielding a 1.7B/40B MoE configuration.

\textbf{Accuracy and inference efficiency evaluation Recipe} Same as Sec.\ref{exp:same_act_inhouse}.

\begin{wrapfigure}{r}{0.5\textwidth} 
    \centering
    \includegraphics[width=0.48\textwidth]{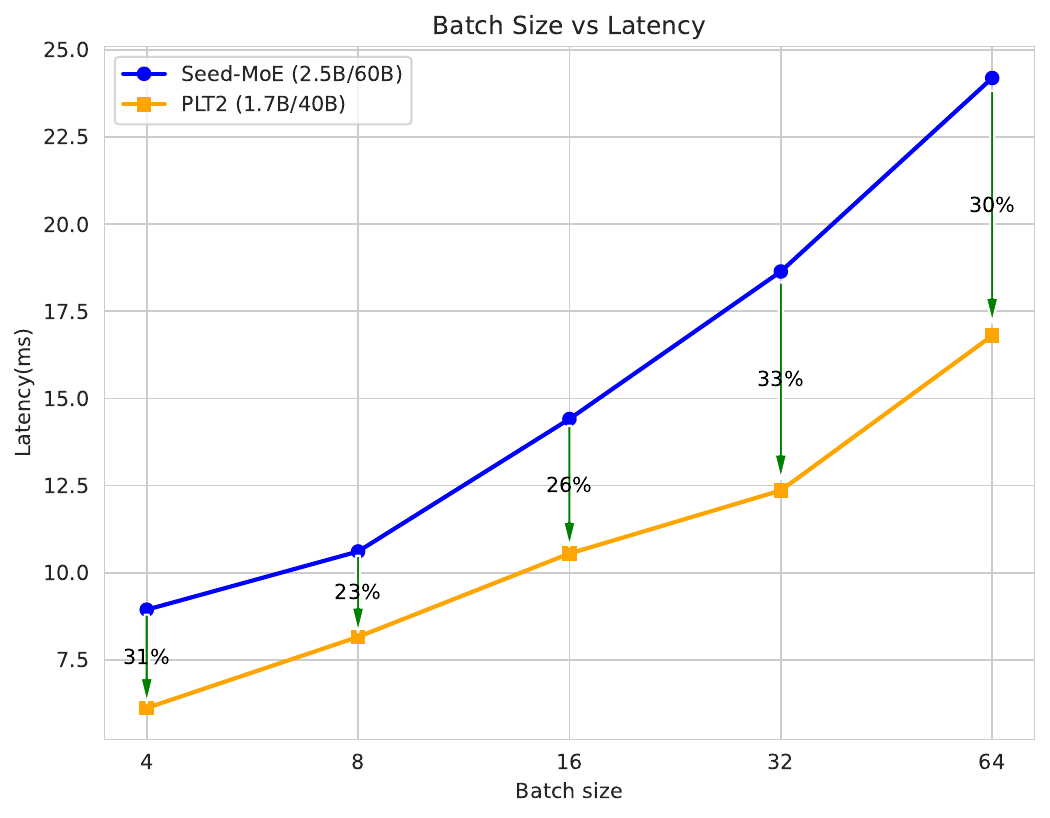}
    \caption{Batch size vs latency on Seed-MoE (2.5B/60B) and PLT-2 (1.7B/40B).}
    \label{fig:inhouse_latency_vs_batch size 2b5}
\end{wrapfigure}

\textbf{Results.} As shown in Table~\ref{tab:seed_2b5_exp}, the 1.7B/40B MoE model with $PLT$ and two loops achieves an average accuracy of $62.6$, outperforming the vanilla 2.5B/60B MoE model by $0.5$ points. Figure~\ref{fig:inhouse_latency_vs_batch size 2b5} illustrates decoding latency across batch sizes from $4$ to $64$. The 1.7B/40B MoE model with $PLT$ and two loops delivers about $30\%$ lower latency than the vanilla 2.5B/60B MoE model. In addition, its KV cache is roughly two-thirds of the baseline due to the reduced depth. Overall, $PLT$ improves the scalability of Transformers, achieving similar accuracy with fewer parameters, lower KV-cache memory, and lower inference latency.

\begin{table}[tp]
    \centering
    \caption{
    Performance comparison of in-house Seed-MoE 2.5B and PLT-2 1.7B. 
    }
    \scriptsize
    \begin{adjustbox}{max width=\textwidth}
    \begin{tabularx}{\textwidth}{l|XXXXXXXXXX|X}
    \toprule
     & MMLU & CEval & M.PRO & AGI. & BBH & MATH & TQA & DROP & MBPP & H.Eval & Avg. \\
    \midrule
    Seed-MoE (2.5B/60B) &  75.1 & 78.2 & \emph{47.4} & \emph{60.6} & \emph{70.9} & \emph{48.6}  & 66.3 & 63.9 & 60.8 & 49.4 & 62.1  \\
    \textit{PLT}-2 (1.7B/40B) &  \emph{77.3} & \emph{80.5} & 45.8 & 58.9 & 66.7 & 43.5 & \emph{71.6} & \emph{65.0} & \emph{64.0} & \emph{52.4} & \emph{62.6}   \\

    \bottomrule
    \end{tabularx}
    \end{adjustbox}
    \label{tab:seed_2b5_exp}
\end{table}

\section{Related Works}


\subsection{Latent Reasoning}

While Chain-of-Thought (CoT)~\citep{wei2022chain} improves reasoning through explicit intermediate step generation, researchers have also explored latent reasoning paradigms that internalize or extend such multi-step computation within the model itself -- either vertically (looped Transformers) or horizontally (latent CoT).

\paragraph{Looped Transformers -- vertical latent reasoning} 
The initial studies by \citet{dehghani2018universal} and \citet{lan2020albert} introduced parameter sharing and recurrent adoption across layers, laying the foundation of looped Transformers. \citet{saunshi2025reasoning} highlighted the strong capabilities of looped Transformers in complex reasoning tasks. Meanwhile, \citet{yang2023looped,fan2024looped} demonstrated its parameter efficiency in data-fitting settings, and superior length generalization on RASP-L tasks. More recent works such as \citet{chen2025inner} and \citet{geiping2025scaling} allocate greater computational depth to complex tokens through looping, enabling models to scale up inference-time reasoning capacity without increasing parameter count.

\paragraph{Latent CoT -- horizontal latent reasoning}
Previous works~\citep{goyalthink,zelikman2024quiet} enhance the model’s reasoning capability by inserting special discrete, non-semantic tokens into the sequence, thereby allocating additional computational steps to refine its intermediate representations.
In contrast, recent studies by~\citep{hao2024training,tack2025llm} compress  reasoning steps into continuous latent thoughts, achieving more efficient and expressive internal reasoning.
These latent CoT methods demonstrate notable improvements across reasoning benchmarks such as GSM8K~\citep{cobbe2021training}, NaturalQuestions~\citep{kwiatkowski2019natural}, and CommonsenseQA~\citep{talmor2018commonsenseqa}, etc.

Both looped Transformers and latent CoT reasoning suffer from inferior inference efficiency due to their inherently sequential computation -- whether across loops or by tokens, while our proposed \method innovatively overlaps the computation of different loops in different tokens, leading to extreme inference efficiency.

\subsection{Parallelized Computation}
The parallelized computation during inference in Large Language Models is a relatively new research area in recent years. We highlight three representative works including PHD~\citep{wu2025efficient}, ParScale~\citep{chen2025parallel}, and StagFormer~\citep{cutler2025stagformer}

\paragraph{PHD} PHD~\cite{wu2025efficient} presents that utilizing parallel computation can leads to scalable performance improvement via parallelizing the forward of repeated tokens. To maintain minimum memory access overhead while achieve better performance, PHD introduces both KV cache sharing and chunk-wise sliding window attention. The drawback of PHD lies in the token repetition methodology, which is an inefficient method of utilizing parallel computation, since the hidden representations in the former transformer layers are very similar. Under high-throughput serving scenarios, the improved performance of PHD can not compensate for the loss of throughput due to increased decoding computation.

\paragraph{ParScale} Latter, ParScale~\cite{chen2025parallel} presents that using sequence repetition with prefix tuning~\cite{li2021prefix} can also leads to scalable performance improvement. The drawback of ParScale lies in its inefficiency by introducing $P\times$ KV cache when $P$ inference streams are activated, leading to overhead both in KV cache footprint and inference latency, especially in the high-throughput serving scenarios.

\paragraph{StagFormer} StagFormer~\citep{cutler2025stagformer} proposes a time-staggered decoding mechanism that parallelizes Transformer layer execution along the depth axis by splitting layers into multiple stacks, where the upper stack attends to the lower stack’s hidden states from the previous time step via cross-attention. This achieves partial layer-level parallelism but still suffers from incomplete parallelism in attention computation and memory access, leading to limited efficiency gains. Specifically, The separate-weight variant doubles hardware usage but achieves less than $50\%$ throughput improvement, while the weight-sharing variant, though lighter in parameters, incurs extra KV-cache cost and additional cross-attention overhead compared to our \textit{loop2+CLP} design, resulting in strictly worse inference efficiency. With completed parallelism and sharing strategy over the looped transformer, our proposed \method further presents more impressive inference efficiency by reducing the KV cache footprint by over $50\%$ with no performance degradation.

In this paper, our presented \method improves the utilization of parallel computation to an unprecedented level. By discovering loop transformers are naturally suitable for parallel computation utilization, we propose cross-loop parallelism to improve looped transformer, maintaining the performance of loop transformers while achieving especially better inference efficiency. 

\section{Conclusion}


In this paper, we present Parallelized Looped Transformer (PLT), which proposes \textbf{Cross-Loop Parallelism (CLP)} that overlaps \emph{the computation and the memory-access} of \textit{latter loops in previous tokens} and \textit{previous loops in latter tokens}, and \textbf{gated sliding window attention (gated-SWA)} to achieve \emph{parallelism in the access of KV cache} with no performance degradation. {PLT} presents impressive performance improvement with negligible latency overhead compared with the vanilla Transformer under memory-access bottle-neck, and clearly better inference efficiency with similar or even better performance compared with vanilla looped transformers with the same inference computation budget, showing the potential of parallel computation utilization.

\newpage
\section{Contributions and Acknowledgments} \label{sec:contribution}
Names are Listed in the alphabetic order.
\subsection*{\textcolor{brown}{Project Lead}}
Bohong Wu
\subsection*{\textcolor{brown}{Core Contributors}}
Mengzhao Chen, Xiang Luo, Bohong Wu, Shen Yan
\subsection*{\textcolor{brown}{Infrastructure}}
Xiang Luo, Fan Xia, Tianqi Zhang, Hongrui Zhan, Zheng Zhong
\subsection*{\textcolor{brown}{Model Architecture}}
Mengzhao Chen, Bohong Wu, Shen Yan, Qifan Yu, Xun Zhou
\subsection*{\textcolor{brown}{Supervision}}
Siyuan Qiao, Xingyan Bin

\subsection*{Acknowledgments}
We thank Jianqiao Lu, Yunshui Li, Yao Luo, Jin Ma, Yiyuan Ma, Yutao Zeng, Chaoyi Zhang, Zhijian Zhuo as well as other colleagues at ByteDance for their support for this project.

\clearpage

\bibliographystyle{plainnat}
\bibliography{main}

\clearpage

\beginappendix
\section{Experiments opensource} \label{exp:opensource}

We arrange our opensource experiments in two-fold.
\begin{itemize}
    \item We present the open source Experiments on both the dense model series and MoE model series, with over 1 billion activated Parameters in Section~\ref{exp:part1}.

    \item We present similar ablation study in Section~\ref{exp:component} on the dense models, as a complementary of the our in-house Seed-MoE experiments.

\end{itemize}

\subsection{Evaluation Datasets}

We used the following open-source datasets in our evaluation, including MMLU~\citep{hendrycks2020measuring}, HellaSwag~\citep{zellers2019hellaswag}, ARC~\citep{clark2018arc}, PIQA~\citep{bisk2020piqa}, Winogrande~\citep{sakaguchi2021winogrande}, CommonsenseQA~\citep{talmor2018commonsenseqa}.

\begin{figure}[tp]
    \centering
    \begin{subfigure}[b]{0.23\textwidth}
        \centering
        \includegraphics[width=\textwidth]{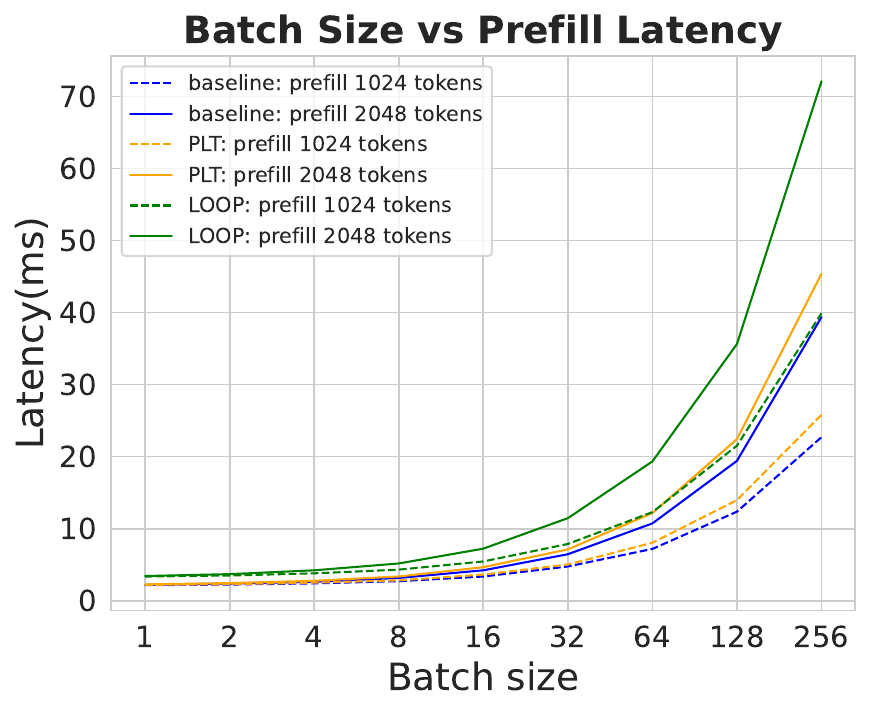}
        \caption{Dense latency.}
        \label{fig:dense_latency}
    \end{subfigure}
    \begin{subfigure}[b]{0.23\textwidth}
        \centering
        \includegraphics[width=\textwidth]{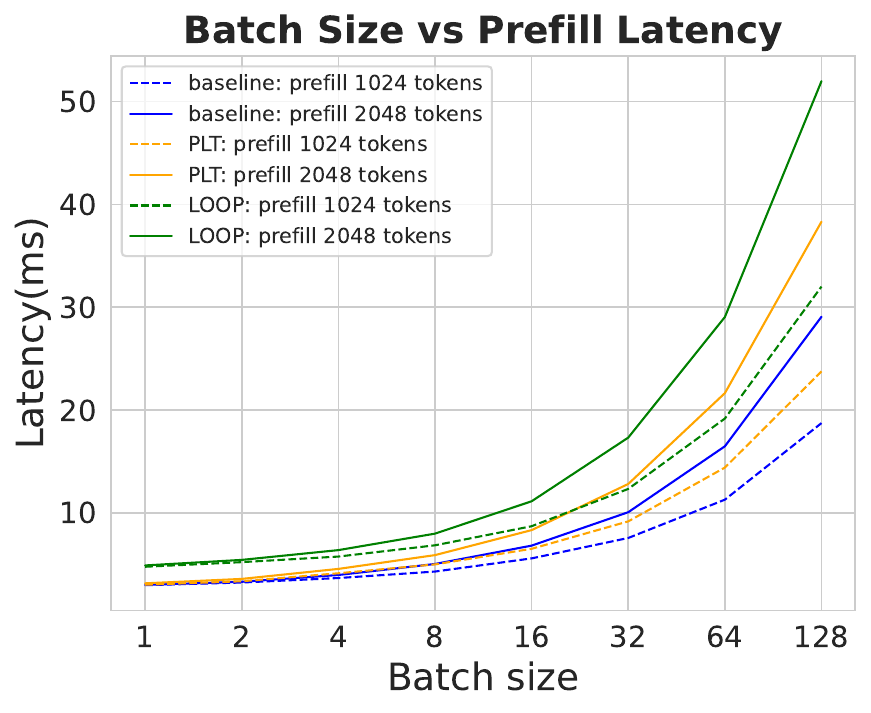}
        \caption{MoE latency.}
        \label{fig:moe_latency}
    \end{subfigure}
    \begin{subfigure}[b]{0.23\textwidth}
        \centering
        \includegraphics[width=\textwidth]{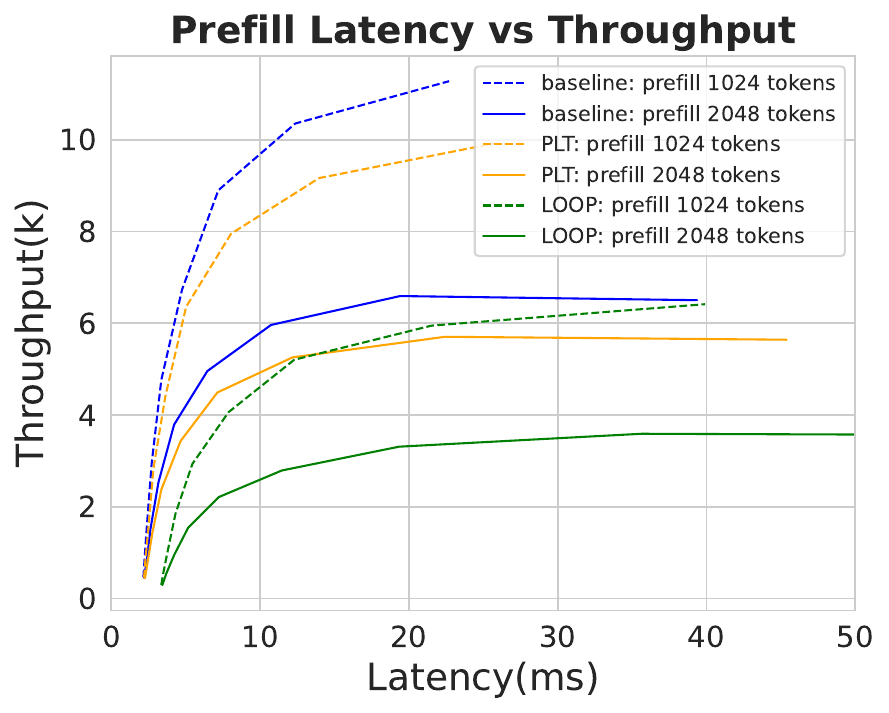}
        \caption{Dense throughput.}
        \label{fig:dense_throughput}
    \end{subfigure}
    \begin{subfigure}[b]{0.23\textwidth}
        \centering
        \includegraphics[width=\textwidth]{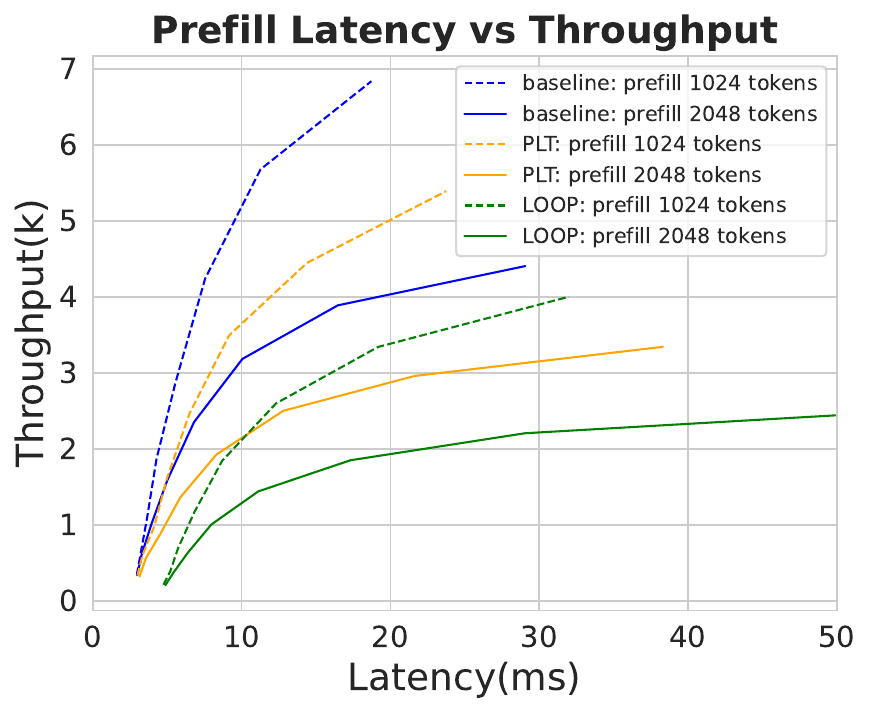}
        \caption{MoE throughput.}
        \label{fig:moe_throughput}
    \end{subfigure}
    \caption{Inference efficiency analysis including latency and throughput for vanilla transformer, \method and looped transformer over 1 billion activated parameters. We use FP8 quantization during inference based on VLLM~\citep{Kwon2023vllm}.}
    \label{fig: phd_opensource_efficiency.}
\end{figure}

\begin{table}[tp]
    \centering
    \small
    \setlength{\tabcolsep}{7pt}
    \caption{
    Performance evaluation of 1.2B dense models and 1B/7B MoE models across the baseline, vanilla looped transformer with $L=2$ and our proposed \method with $L=2$. 
    Evaluated benchmarks include: MMLU, Hellaswag (Hella.), ARC-Challenge (ARC-C), ARC-Easy (ARC-E), PIQA, Winogrande (Wino.), and Commonsense QA (Comm.).
    }
    \begin{tabular}{l|c|ccccccc|c}
    \toprule
    & Loss & MMLU & Hella. & ARC-C & ARC-E & PIQA & Wino. & Comm. & Avg. \\
    \midrule
        Vanilla Dense & 2.577 & 35.5 & 62.5 & 38.1 & 71.4 & 74.9 & 60.6 & 44.7 & 55.4 \\
        \ \ + \textit{loop-2} & \textbf{2.532} & 36.3 & \textbf{66.1} & 40.5 & \textbf{73.9} & 75.6 & \textbf{64.6} & 46.4 & \textbf{57.6} \\
        \ \ + \textit{PLT-2} & 2.537 & \textbf{36.8} & 65.4 & \textbf{41.5} & 72.3 & \textbf{76.5} & 61.4 & \textbf{47.9} & 57.4 \\
    
    \midrule
        Vanilla MoE & 2.342 & 37.3 & 67.2 & 40.5 & 72.1 & 76.3 & 62.7 & 48.2 & 57.8 \\
        \ \ + \textit{loop-2} & 2.302 & \textbf{38.7} & 70.3 & \textbf{44.1} & 75.1 & 77.0 & \textbf{64.1 }& 48.2 & 59.6 \\
        \ \ + \textit{PLT-2} & \textbf{2.280} & 38.2 & \textbf{71.0} & 43.1 & \textbf{77.2} & \textbf{78.7} & 63.5 & \textbf{48.7} & \textbf{60.0} \\
        
    \bottomrule
    \end{tabular}
    \label{tab:main_1b2}
\end{table}

\begin{table}[tp]
    \centering
    \small
    \setlength{\tabcolsep}{2pt}
    \caption{Component analysis of \method. Compared with vanilla looped transformer, we introduce two extra components for the consideration of both inference speed and performance. $Lat.$ is the abbreviation for latency(ms).}
    \begin{tabular}{l|c|ccccc|c|cc}
    \toprule
    & Loss & MMLU & Hella. & ARC-C & ARC-E & PIQA & Avg. & Lat.$_{@bs=1}$ & Lat.$_{@bs=16}$ \\
    \midrule
        (1) Vanilla Dense & 2.880 & 29.2 & 45.7 & 26.4 & 61.8 & 69.7 & 46.6 & \colorbox{lightgreen}{1.68} & \colorbox{lightgreen}{2.49} \\
        (2) +\textit{loop-2} & 2.856 & 28.7 & 47.1 & 26.1 & 60.9 & \textbf{70.6}  & 46.7 & \colorbox{lightred}{2.66} & \colorbox{lightred}{3.92} \\
        (3) +\textit{loop-2}+${CLP}$ & \textbf{2.840} & \textbf{30.0} & \textbf{48.3} & \textbf{30.8} & 59.5 & 69.7 &  \textbf{47.8} & \colorbox{lightgreen}{1.73} & \colorbox{lightyellow}{3.23} \\
        (4) +\textit{loop-2}+${CLP}$+${KV share}$ & 2.858 & 29.6 & 46.5 & 28.4 & 60.2 & 69.8 &  46.9 & \colorbox{lightgreen}{1.73} & \colorbox{lightgreen}{2.69} \\
        (5) +\textit{loop-2}+${CLP}$+${KV share}$+${G{-}SWA}$ & 2.844 & 29.7 & 47.4 & 30.1 & \textbf{62.3} & 69.6 &  \textbf{47.8} & \colorbox{lightgreen}{1.73} & \colorbox{lightgreen}{2.70} \\

    \bottomrule
    \end{tabular}
    \label{tab: component_analysis}
\end{table}

\subsection{Main Experiments} \label{exp:part1}

\subsubsection{Training and Evaluation Settings} \label{sec:training setting}
Our open-source implementation is based on OLMo and OLMoE. We compare our method with the vanilla transformer and vanilla looped transformer with the same activated parameters. 

\paragraph{Dense Models Recipe} 
For dense models, we set the number of transformer layers to 16. We set the hidden dimension to 2048. 
For MLP modules in these models, we set the MLP hidden dimension to 16384. We use FFN output norm.
For Attention modules in these models, we use GQA with 32 query heads and 8 key/value heads. We use query/key/value layernorm at the same time, while no attention output norm.
For LM head modules, we use weight tying that shares the parameter with the embedding layer.

For training dynamics, we train the model for 400B tokens in total, with global training batch size set to 1024. We use the cosine learning rate schedule, using 3e-4 as the peak learning rate and 2000 steps warmup.

Note that compared with the original setting of public available opensource model OLMo-1B, we have made slight modifications on the norm settings due to the consideration of training stability~\citep{zhuo2025hybridnorm}.

\paragraph{MoE Models Recipe}
For MoE models, we set the number of transformer layers to 16. We set the hidden dimension to 2048.
For MLP modules in these models, we use SwiGLU experts and the 8 in 64 recipe. 
For Attention modules in these models, we use MHA with 16 attention heads in total. 
For LM head modules, we also use weight tying that shares the parameter with the embedding layer.

For training dynamics, we train the model for 400B tokens in total, with global training batch size also set to 1024. We use the cosine learning rate schedule, using 4e-4 as the peak learning rate and 2500 steps warmup.

\paragraph{Efficiency Evaluation Recipe} Based on VLLM~\citep{Kwon2023vllm}, we analyze the efficiency including both the latency and throughput, of all these baseline models. We vary the prefilling length in $[1024, 2048]$ and the serving batch size in $[1, 2, 4, 8, 16, 32, 64, 128, 256]$, which gradually shifts from memory-access bottleneck to compute bottleneck. For serving, we use FP8 quantization which is closer to the real industrial serving scenarios. The latency metric is averaged across 5 independent runs of decoding 256 tokens. We conduct the efficiency analysis experiments on one single GPU.

\subsubsection{Analysis}

Table~\ref{tab:main_1b2} presents the performance of variants on Dense Models and MoE models. The observations are similar with our in-house experiments that \emph{PLT-2 achieves very similar performance with vanilla looped transformers}. Figure~\ref{fig: phd_opensource_efficiency.} further presents the efficiency analysis, where we also obtain similar observations that \emph{PLT-2 presents obviously better efficiency than vanilla looped transformers}.


\subsection{Ablation study} \label{exp:component}

\subsubsection{Training/Evaluation Settings} \label{train_setting:abl}
\paragraph{Dense Models Recipe}
The training settings in this section mainly follows the setting in Section~\ref{sec:training setting}. Except for the following two modifications. 
\begin{itemize}
    \item We change GQA with 16 query heads and 4 key value heads, and set hidden dimension to 1536 and MLP hidden size to 8192.
    \item We change the training token numbers to 100B, with 3e-4 as the peak learning rate and the cosine learning rate scheduler, warm-up 2000 steps.
\end{itemize}

\paragraph{MoE Models Recipe}
The training settings in this section mainly follows the setting in Section~\ref{sec:training setting}. Different recipes are also presented as follows.

\begin{itemize}
    \item We change the layer numbers to 12 and hidden dimension to 1536.
    \item For training dynamics, we also set the number of training tokens to 100B, with 3e-4 as the peak learning rate and the cosine learning rate scheduler, warm-up 2000 steps.
\end{itemize}

\paragraph{Evaluation Settings} \label{ablation_study:eval}
The evaluation settings mainly follows Section~\ref{sec:training setting}, except for we use one H800 GPU for evaluation.

\subsubsection{Analysis}

Table~\ref{ablation_study:eval} presents our ablation study on OLMo2, where the observation are similar with our in-house experiments.

\end{document}